\documentclass[conference]{IEEEtran}
\IEEEoverridecommandlockouts
\usepackage{cite}
\usepackage{amsmath,amssymb,amsfonts}
\usepackage{algorithmic}
\usepackage{graphicx}
\usepackage{textcomp}
\usepackage{xcolor}
\def\BibTeX{{\rm B\kern-.05em{\sc i\kern-.025em b}\kern-.08em
    T\kern-.1667em\lower.7ex\hbox{E}\kern-.125emX}}
    
\usepackage{flushend}
\usepackage{balance}
\begin{document}

\title{Is Simple Better? Revisiting Non-linear Matrix Factorization for Learning Incomplete Ratings\\
}

\author{
\IEEEauthorblockN{ Vaibhav Krishna}
\IEEEauthorblockA{\textit{ETH Zurich} \\
Zurich, Switzerland \\
vaibhavkrishna@ethz.ch}
\and
\IEEEauthorblockN{ Tian Guo}
\IEEEauthorblockA{\textit{ETH Zurich} \\
Zurich, Switzerland \\
tian.guo@gess.ethz.ch}
\and
\IEEEauthorblockN{ Nino Antulov-Fantulin}
\IEEEauthorblockA{\textit{ETH Zurich} \\
Zurich, Switzerland \\
anino@ethz.ch}
}

\maketitle

\begin{abstract}
Matrix factorization techniques have been widely used as a method for collaborative filtering for recommender systems. In recent times, different variants of deep learning algorithms have been explored in this setting to improve the task of making a personalized recommendation with user-item interaction data. The idea that the mapping between the latent user or item factors and the original features is highly nonlinear suggest that classical matrix factorization techniques are no longer sufficient. In this paper, we propose a multilayer nonlinear semi-nonnegative matrix factorization method, with the motivation that user-item interactions can be modeled more accurately using a linear combination of non-linear item features. Firstly, we learn latent factors for representations of users and items from the designed multilayer nonlinear Semi-NMF approach using explicit ratings. Secondly, the architecture built is compared with deep-learning algorithms like Restricted Boltzmann Machine and state-of-the-art Deep Matrix factorization techniques. 
By using both supervised rate prediction task and unsupervised clustering in latent item space, we demonstrate that our proposed approach achieves better generalization ability in prediction as well as comparable representation ability as deep matrix factorization in the clustering task. 
\end{abstract}


\section{Introduction}
Nowadays, recommender systems \cite{Schafer2001, Sarwar2001, Bobadilla2013,Smyth2007} are so ubiquitous in information systems that their absence draws attention and not vice versa.
Making personalized predictions for specific users, based on some functional dependency of past interactions of all users and items is known as collaborative filtering \cite{Su2009,Adomavicius2012} approach. 
Within this approach, different matrix factorization (MF) techniques have been proven to be quite accurate and scalable for many recommender scenarios \cite{Koren2009,Koren2011, XinLuo2014}. Essentially, they map both users and items to a joint latent factor space of lower dimension and model interaction as their inner product in latent space.

Recently, various deep learning models \cite{Bengio2013}
such as Restricted Boltzmann Machines \cite{Salakhutdinov2007}, stacked auto-encoders \cite{Wang2015,Li2015}, deep neural networks \cite{He2017} or deep matrix factorizations \cite{Xue2017} were introduced to the collaborative filtering methods. Two aspects of collaborative filtering were upgraded: (i) linear latent representations of users and items were replaced by deep representations and (ii) inner product was replaced by non-linear function represented with deep neural networks. 
This has motivated us to study which of this two aspects: (i) non-linear latent representations or (ii) non-linear interaction functions dominates the ability to learn incomplete ratings. 
In this paper, we are focused on the matrix factorization recommender models. Note, that the representations of classical matrix factorizations such as non-negative matrix factorization (NMF) or singular value decomposition (SVD) are essentially the same as the ones learned by a basic linear auto-encoders \cite{Bengio2013}. 


In this paper, we focus on the collaborative filtering task of learning explicit ratings based on collaborative filtering.
For recent advances in using deep models with an auxiliary information and implicit feedback  \cite{Karatzoglou2017}.
The main contributions of the paper are the following: 
(i) We propose a simple model of the compositions of non-linear matrix factors for learning incomplete explicit ratings. 
(ii) We evaluated our approach against a variety of baselines including both linear and non-linear methods. 
(iii) In the supervised rate prediction task, our simple linear combination of non-linear representations has lower prediction errors (RMSE) on hold-out datasets, thereby leading to better generalization ability. 
(iv) In the unsupervised clustering task performed on the obtained representations, we demonstrate that our approach attains comparable representation ability in comparison to complex deep matrix factorization, by presenting comparable clustering performance metric (within-cluster sum of squares). 

\section{Preliminaries}
In this section, we formulate the problem setting. 
The user-item rating matrix is denoted by $R\in \mathcal{R}^{n \times m}$. 
The original representation of user $i$ is the i-th row of matrix $R$ i.e. $u_i \in \mathcal{R}^{m}$, while the original representation of item $j$ is the $j$-th column of matrix $R$ i.e.  $v_j \in \mathcal{R}^{n}$.
Let $P\in \mathcal{R}^{n \times k }$ denote the latent feature matrix of users, where $p_i$ denotes the latent feature vector of user $i$ i.e. $i$-th row of matrix $P$. 
Similarly, let $Q \in \mathcal{R}^{k \times m}$ denote the latent feature matrix of items, where $q_j$ denotes the latent feature vector of item $j$ i.e. $j$-th column of $Q$. 

The collaborative filtering task is to estimate the unobserved rating for user $i$ and item $j$ as
\begin{equation}
\hat r_{i,j} = f(p_i, q_j | \Theta),
\end{equation}
where $\Theta$ denotes the model parameters of interaction function $f$. 

The latent features $p_i, q_j$ and parameters $\Theta$ are found by minimizing the objective function
\begin{equation}
\label{eq:objective}
L = \sum_{i,j \in \kappa} \mathcal{L}(r_{ij}, \hat r_{ij}) + \lambda \Omega(\Theta),
\end{equation}
where $\mathcal{L}(.)$ is a point-wise loss function (for pair-wise learning see \cite{Cao2007}), $\Omega(.)$ is a regularizer (L2 or L1 norm \cite{Ning2011}) and $\kappa$ denotes sets of training instances over which learning is done (see \cite{Steck2010}, about the missing data assumptions). 


The regularized matrix factorization \cite{Koren2009, XinLuo2014} approach is modeling the interaction function $f$ as the linear combination $f(p_i, q_j | \Theta)=\sum_k p_i(k) q_j(k)$, which represent the inner product in the latent space. The latent factors $p_i, q_j$ are linear representations, due to the absence of non-linearity in the linear algebra transformations $R \approx PQ$. Here, usually L2 function is used for loss and regularization, while for the training set $\kappa$, set of observed ratings were used.

Neural collaborative filtering \cite{He2017} models the interaction function $f$ as a deep neural network $f(p_i, q_j | \Theta) = \phi_{out}(\phi_x(...(\phi_1(p_i, q_j ))...))$, where $\phi_{out}$ and $\phi_{x}$ are mapping of output layer and x-th layer of the form $\phi(z)=\sigma(Wz+b)$. The $\sigma(.)$ represents the non-linear function such as rectified linear unit, hyperbolic tangent or others.  The model parameters $\Theta$ and latent factors $p_i, q_j$ are learned in a joint manner, where the latent factors are non-linear representations of user and item vectors $u_i,v_j$ from rating matrix $R$. 

Deep matrix factorizations \cite{Xue2017} model the interaction function $f$ as an inner product $f(p_i, q_j | \Theta)=\sum_k p_i(k) q_j(k)$. However, both user and item features are deep representations of the form: $p_i=\phi_{out}(\phi_x(...(\phi_1(u_i))...))$ and $q_j=\phi_{out}(\phi_x(...(\phi_1(v_j))...))$.

\section{Proposed Model}
In order to learn incomplete ratings, we formulate the following multilayer semi-NMF model (NSNMF):
$R^ \approx B + P_1 Q_1^{+}$,
where $R \in R^{n\times m}$ is the rating matrix, $B\in R^{n\times m}$ is the bias matrix, $P_1 \in R^{n\times k}$ is the latent user preferences matrix and $Q_1^{+} \in R_{+}^{k\times m}$ is the matrix of the non-linear item latent features. To model non-linear representation of the items, we use the following model  $Q_1^+ \approx g(S_{2} Q_{2}^+)$, which finally leads to: \\ 
\begin{equation}
\label{eq:mainModel}
R \approx B + P_1 g(S_2 Q_2^{+}),
\end{equation}
where $Q_2^{+} \in R_{+}^{l\times m}$ is the hidden latent representation of items, $S_2$ is the weighting matrix between latent representations on different levels and $g(.)$ is the element-wise non-linear function to better approximate the non-linear manifolds of the latent features.

Similar architecture to our proposed method without bias term is used in the task of learning deep representation of images \cite{Trigeorgis2017}, where the learning depends on dense multiplicative updates.

Simply, we model the interaction function $f$ as an inner product with offset $f(p_i, q_j | \Theta)= b_{i,j}+\sum_k p_i(k) \cdot q_j(k)$, where user feature $p_i$ is the $i$-th row of matrix $P_1$ and item non-linear representation $q_j$ is the $j$-th column of 
$Q_1^+ \approx g(S_{2} Q_{2}^+)$. 
This model falls into the category of semi-NMF factorization models \cite{Ding2010}. Semi-NMF is a variant of NMF, which imposes non-negativity constraints only on the latent factors of the second layer $Q^{+}$. This allows both positive and negative offsets from the bias term $B$. 
Usually, in clustering $P$ represents cluster centroids and $Q$ represents soft membership indicator \cite{Ding2010}.
But, from recommender perspective, matrix $P$ may be interpreted as the linear regression coefficients and the nonnegativity constraints imposed on latent item attributes $Q$ allow part-based representations \cite{Seung1999,Seung2001}.

Note, that much of the observed variation in rating values is due to the effects associated with either users or items, known as biases or intercepts, independent of any interactions. Thus, instead of explaining the full rating value by an interaction of the form $p_i.q_j^T$, the system can try to identify the portion of these values that individual user or item biases cannot explain.
Bias $b_{ui}$ involved in a rating $r_{ui}$ can be computed as follows:
$ b_{ui} = \mu + b_u + b_i$,
where
$\mu$ \textit{is the overall average rating},
$b_u$ \textit{is the observed deviations of user u} and
$b_i$ \textit{is the observed deviations of item i}.

Note, that in general case, we are able to compose more non-linearities with the following relation
$Q_j^+ \approx g(S_{j+1} Q_{j+1}^+)$,
which generates the following model
$R \approx B + P_1 g(S_2 g(S_3 g(... g(S_f Q_f^{+}))))$.
However, more complex item latent representation $f \geq 3$ were showed not to be useful, see experiments section for more details. 





\subsection{Learning Model Parameters} 
For a two layered item features structure, the model parameters are updated through an element-wise gradient descent approach, minimizing eq.(\ref{eq:objective}), with squared loss function
$\mathcal{L}(r_{ij}, \hat r_{ij}) = (r_{ui} -\hat r_{ui})^2$ and L2 regularization $\Omega(\Theta)= b_u^2 + b_i^2 + \| p_u \|^2 + \| q_i \|^2 + \| s_{ui} \|^2$.

The model parameters $\Theta$ are randomly initialized uniformly at the range [0,1], and we perform iterative updates for each observed rating $r_{ui}$ as follows:

\begin{equation}
b^*_{u} \gets b_{u} + \eta (e_{ui} - \lambda b_u)
\end{equation}
\begin{equation}
b^*_{i} \gets b_{i} + \eta (e_{ui} - \lambda b_i)
\end{equation}
\begin{equation}
p^*_{uk} \gets p_{uk} + \eta \left(e_{ui} \cdot g\left(\sum_{h}s_{kh} \cdot q_{hi}\right) - \lambda p_{uk}\right)
\end{equation}
\begin{equation}
s^*_{kl} \gets s_{kl} + \eta \left(e_{ui}\cdot p_{uk} \cdot g'\left(\sum_{h}s_{kh} \cdot q_{hi}\right)q_{li} - \lambda s_{kl}\right),
\end{equation}
\hspace{100pt} \textit{(update only if $g(S^*_{[k.]} Q^*_{[.i]}) > 0) $}
\begin{equation}
q^*_{li} \gets q_{li} + \eta \left(e_{ui} \cdot p_{uk} \cdot g'\left(\sum_{h}s_{kh} \cdot q_{hi}\right)s_{kl}- \lambda q_{li}\right),
\end{equation}
\hspace{100pt} \textit{(update only if $q^*_{li} > 0) $},
\\
where \textit{g'(.)} is the derivative of activation function and $e_{ui} =  r_{ui} - \hat r_{ui}$ is the error term. 
Note, that we do not explicitly store the dense matrix $B$. 
The computational complexity for training a 2-layer item-feature NSNMF architecture is of order $\mathcal{O}(t(m+n)(kl+kl^2))$, where \textit{k,l} are the dimensions of layer $S_2$, and \textit{t} the number of iterations. The learning rate $\eta$ was configured with the AdaGrad method \cite{AdaGrad}, performing larger updates for infrequent and smaller updates for frequent parameters. For this reason, it is well-suited for dealing with sparse data, as in our case of incomplete ratings. 
Given \textit{k,l} are constant and $\textit{(m+n)} \gg \textit{k,l}$, the scalability of the proposed method linearly depend on the number of users and items.



\section{Experimental evaluation}
In this section, we report experimental results by evaluating our NSNMF approach and a variety of baselines in both supervised and unsupervised tasks. 

\subsection{Datasets}
We have three real datasets as follows: 
\begin{itemize}
\item MovieLens100K \footnote{https://grouplens.org/datasets/movielens/100k/, generated on October 17, 2016.}
- 100004 ratings, across 9125 movies from 671 users. (have users with at least 20 ratings) 
\item FilmTrust 
\footnote{https://www.librec.net/datasets.html}
- 28496 ratings, across 1981 movies from 654 users. (filtered to have users with at least 20 ratings)
\item Amazon Music 
\footnote{http://jmcauley.ucsd.edu/data/amazon/}
- 50395 ratings, across 1188 items from 19260 users. (filtered to have users with at least 20 ratings and items with at least 2 interactions)
\end{itemize}

Each dataset is split into training ($80\%$), and testing sets ($20\%$). The training set is then used for 10-fold cross-validation for hyperparameter tuning. 


\subsection{Baselines and setup}
We use baselines including both linear and nonlinear approaches as follows:

\textbf{CF} 
Neighborhood models are the most common approach to CF with user-oriented and item-oriented methods \cite{Sarwar2001,Karypis2001}. They are respectively referred to as \textbf{User-User CF} and \textbf{Item-Item CF}.

\textbf{SVD} is applied in the collaborative filtering domain by factorizing the user-item rating matrix \cite{sarwar2000application} by updating only for the known ratings. 

\textbf{NMF} 
Nonnegative matrix factorization (NMF) which introduces the nonnegative constraint into a MF process \cite{XinLuo2014,zhang2006learning} along with the \textbf{regularized NMF} mode to avoid over-fitting.

\textbf{RBM} 
Restricted Boltzmann Machine (RBM) \cite{Salakhutdinov2007} is an undirected graphical model, which contains a layer of visible softmax units for items and a hidden binary unit for user rating.
Each hidden unit could then learn to model a significant dependency between the ratings of different movies.

\textbf{DMF} presents a deep structure learning architecture to learn deep low dimensional representations respectively for users and items \cite{Xue2017}.
They use both explicit ratings and implicit feedback for a better optimization of a normalized cross entropy loss function to predict scaled ratings on a continuous scale [0,1]. 


As for \textbf{our proposed NSNMF} method, we evaluate it with different activation functions as follows: (i) \textbf{NSNMF ReLU} is the NSNMF model with Rectified Linear unit: $\text{ReLU}(x) = \max(x,0) $ as the activation function, (ii) \textbf{NSNMF SoftPlus} uses softplus: $ \text{softplus}(x) = \log(1+e^x)$  as the activation function and (iii) \textbf{NSNMF ReLU\_bias} is the proposed model with rectified linear unit activation function plus bias.

In the supervised task, since we focus on explicit ratings,
the rooted mean square error is used to assess the rate prediction performance \cite{Herlocker2004}:
$RMSE = (\frac{1}{N} \sum\limits_{(u,i)}( r_{ui} - \hat r_{ui} )^2 )^{0.5}$.
The less the value of RMSE, the better the approach performs. 

In the unsupervised task, 
we aim to inspect the difference in representations obtained by our NSNMF and baseline approaches. We choose an unsupervised clustering task on such representations and thus
use the pooled within-cluster sum of squares around the cluster means (WCSS) \cite{Tibshirani2001}:
$WCSS = \sum_{r=1}^k \frac{1}{2n_r}\sum_{i,j \in C_r} d_{i,j}$,
where $n_r$ denotes the number of elements inside of cluster $C_r$ and $d_{i,j}$ is the euclidean distance between instances $i$ and $j$ within the same cluster.

\subsection{Supervised tasks}
In the supervised task, we perform 10-fold cross-validation error for each dataset to determine the dimensions of the hidden representation and regularizing parameter for each approach. The dimensions of hidden representation were determined from cross-validation result for values in \{4,6,8,10,15,20\}. The learning rate and regularizer value were varied in the range of \{0.1, 0.01, 0.001\}.

The final models were trained with learning rate 0.01 and regularizing parameter 0.1 and factors value of 4,6,8 for FilmTrust, AMusic and MovieLens datasets respectively.

Then, we report the rate prediction RMSE errors in Table \ref{tab:rmse}.


In Table \ref{tab:rmse}, we observe that NSNMF based approaches, i.e. NSNMF ReLU, Softplus and ReLU\_bias outperformed baselines across all datasets. Especially, NSNMF with ReLU\_bias performed the best and achieved up to $20\%$ less RMSE. Meanwhile, DMF which learns non-linear representations has lower RMSE than the baselines based on linear transformation i.e.
User-User CF, Item-Item CF, SVD, NMF and regularized NMF in most of the time.
We trained
\footnote{https://github.com/RuidongZ/Deep\_Matrix\_Factorizatio\_Models}
DMF \cite{Xue2017} using normalized cross entropy loss on both implicit and explicit ratings. The predicted ratings $\hat{R}$ on the scale [0,1], when scaled to original scale [0,max(R)], where max(R) denotes the max score in all ratings, perform worse than baselines compared to real ratings on the same scale with RMSE measure. 
Thus, we use their DMF architecture, trained using squared loss function to predict unscaled ratings, which are then evaluated again with RMSE. 



\begin{table}[h]
\caption{ Test RMSE}
\label{tab:rmse}
\begin{center}
\begin{tabular}{|c|c|c|c|}
\hline
Algorithm & FilmTrust & ML100K & AMusic\\
\hline
User-User CF & 0.963 & 1.005 & 1.011\\
\hline
Item-Item CF & 0.822 & 1.001 & 0.934\\
\hline
SVD & 1.006 & 1.018 & 2.024\\
\hline
NMF & 0.845 & 0.954 & 1.001\\
\hline
Regularized NMF & 0.840 & 0.937 & 0.975\\
\hline
RBM  & 0.918 & 1.008 & 1.104\\
\hline
DMF & 0.821 & 0.948 & 0.946\\
\hline
NSNMF ReLU  & 0.816 & 0.904 & 0.889\\

NSNMF Softplus  & 0.804 & 0.896 & 0.871\\

NSNMF ReLU\_bias & \textbf{0.788} & \textbf{0.887} & \textbf{0.836}\\
\hline
\end{tabular}
\end{center}
\end{table}



Furthermore, we trained our model with different numbers of hidden layers to assess the prediction errors on all three datasets. We found that 2-layer architecture better models the variation in the rating matrix, while deeper layers even decrease the performance. Due to the page limitation, we report the results up to 3 layers in Table \ref{tab:test_deep}. 




\begin{table}[h]
\caption{Test RMSE of NSNMF w.r.t. different number of layers  }
\label{tab:test_deep}
\begin{center}
\begin{tabular}{|c|c|c|c|}
\hline
Algorithm & FilmTrust & ML100K & AMusic\\
\hline
ReLu\_2 layer  & \textbf{0.816} & \textbf{0.904} & \textbf{0.889}\\
\hline
ReLu\_3 layer & 0.842 & {0.938} & 0.932\\

\hline
\end{tabular}
\end{center}
\end{table}







\subsection{Unsupervised task}
In this part, we perform an unsupervised K-means clustering method 
to evaluate the item representation learned by different approaches in latent spaces.
We performed each approach with the hyperparameter set via cross-validation and then obtain the derived representation.



\begin{figure}[!h]
      \centering
      \includegraphics[scale=0.33]{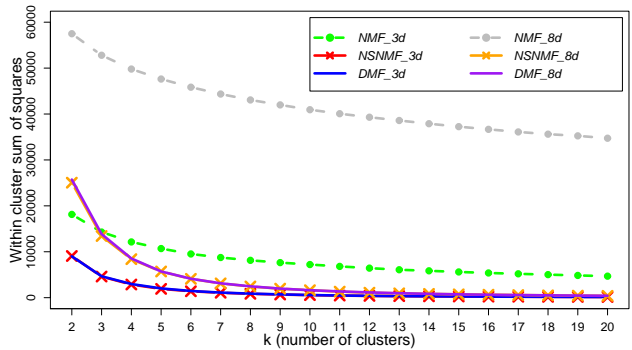}

      \caption{The pooled within-cluster sum of squares around the cluster means (WCSS)  of  clustering on MovieLens100k Dataset (\textit{where 3d, 8d denotes the dimension of feature space})}
      \label{fig:cluster}
\end{figure}

In Figure \ref{fig:cluster}, we report the WCSS \cite{Tibshirani2001}  of each approach w.r.t. the number of clusters. 
We observe that our NSNMF ReLu and DMF constantly yield lower WCSS than NMF. 
It suggests that representation derived by non-linear matrix factorization demonstrates higher representation ability. 
The WCSS of NSNMF ReLu and DMF are quite comparable. It indicates that non-linear transformation is the dominant part while the way of the combination of such representation results in a minor difference in the derived representation. Moreover, the simple linear combination of non-linear representation leads to better generalization ability in supervised prediction, which is already demonstrated in Table \ref{tab:rmse}.  

\section{Discussion and Future Work}

In this paper, 
we focus on learning the non-linear item representations for the explicit feedback and left the extension of learning non-linear implicit feedback representations for future work.
Most of the deep learning architectures have been implemented using dense implicit feedback rating matrix. In this paper, we implement the proposed architecture for explicit feedback only, and left the implementation on implicit feedback for future work. We believe it will be interesting to see the performance of the proposed algorithm on implicit feedback, which will provide a better comparison with the deep learning algorithms \cite{He2017,Wang2015} that are training only on implicit feedback. Thus, the current paper includes the deep learning methods that use explicit feedback in their training algorithms\cite{Salakhutdinov2007,Xue2017}.

We find out that simple linear regression over non-linear item representations is sufficient to overcome the performance of other deep learning methods that use explicit feedback in their training algorithms \cite{Salakhutdinov2007, Xue2017}. It is important to stress, that in our model the linear regression and non-linear item representations are learned in a joint manner via non-linear semi non-negative matrix factorization. The non-negative constraint allows better interpretability of item features e.g. movie cannot have a negative number of certain actors, a negative indication to certain genre etc. However, the semi non-negativity constraint allows the regression coefficients to become negative e.g. negative relation to certain item features.

Furthermore, the linear interaction of non-linear item features provides better predictions than the combination of non-liner item and non-linear user features, as in the case of Deep Matrix Factorization model \cite{Xue2017}.

\section{Conclusions}
We introduced a multilayer nonlinear semi-nonnegative matrix factorization method to learn from the incomplete rating matrix. 
The multilayer approach, which automatically learns the hierarchy of attributes of the items, as well as the non-negative constraint help in the better interpretation of these factors. 
Furthermore, we presented an algorithm for optimizing the factors of our architecture with different non-linearities.
We evaluate our approach in comparison to a variety of matrix factorization and deep learning baselines using both supervised rate prediction and unsupervised clustering in latent item space.
The results offer the insights as follows:
(i) simple linear combination of non-linear representations realized in our proposed approach achieves better generalization ability, that is, lower errors in the prediction on hold-out datasets.
(ii) in the unsupervised clustering task, we find out that the representations learned by our approach yield comparable clustering performance metric (within-cluster sum of squares) as deep matrix factorization. 

\subsection*{Acknowledgement}
N.A.-F. and T.G. are grateful for financial support from the EU Horizon 2020 project SoBigData under grant agreement No. 654024.
The authors acknowledge D. Tolic for useful directions regarding Deep Semi-NMF approach in the early stage of work.

\balance
\bibliography{sample-bibliography}

\vspace{12pt}

\end{document}